# Mechanistic Interpretability of Mamba's Selective Gate in Financial Markets: A Leakage-Controlled Study

Dip Roy[1*], Rajiv Misra[1], Sanjay Kumar Singh[2], Anisha Roy[3]

[1]*Department of Computer Science and Engineering, Indian Institute of Technology, Patna, India*

[2]*Department of Computer Science, Rajarshi School of Management Technology, Varanasi, India*

[3] *Department of Electronics and Communication Engineering, Jaypee Institute of Information Technology, Noida*

*Corresponding author: dip_25s21res37@iitp.ac.in

**Abstract—** Selectively gated state-space models (SSMs), notably Mamba, are increasingly used for financial time series, yet how they internally represent market regimes (bull, bear, sideways) remains opaque. This paper applies Mechanistic Interpretability (MI) to a six-layer Mamba trained on approximately twenty-one years of daily data for a point-in-time set of S&P 500 constituents plus macroeconomic and Fama-French factors, decomposing per-layer hidden states with Top-K Sparse Autoencoders (SAEs). Our central contribution is methodological and cautionary. A prior, uncontrolled configuration recovered market regimes from SAE features at F1 = 1.00. Under a controlled protocol that (i) removes the three input features that overlap the HMM label definition, (ii) disables the auxiliary regime-classification head, (iii) uses a point-in-time constituent universe, and (iv) evaluates against labels disjoint from the inputs, the same regime label is recovered at only F1 = 0.398, indistinguishable from a logistic baseline on raw inputs (0.435) and from an SAE trained on a randomly initialized, untrained Mamba (0.364). A 2x2 attribution grid shows the near-perfect result is an interaction of auxiliary supervision and label-defining inputs: neither factor alone exceeds F1 ~ 0.40. Effect sizes with block-bootstrap inference show the residual selective-gate (Delta t) regime effect is statistically detectable but economically negligible and uniform across layers rather than layer-specialized; the apparent polysemanticity gradient disappears under continuous, theory-grounded measures. We contribute a reproducible leakage-control protocol for interpretability evaluation of financial SSMs, a quantified demonstration that previously reported regime structure is a label-contamination artifact, and a modest economically grounded feature subset that genuinely survives the controls.

## 1. Introduction

Deep Learning has dramatically changed Quantitative Finance on a commercial basis. Algorithmic Trading is responsible for approximately 70% of U.S. Equity Volume [1], and recently there have been numerous advancements in developing more sophisticated Neural Architectures. In addition to these rapid developments in the industry,

Selective State Space Models (SSMs) are gaining popularity as viable computational-efficient alternatives to Transformer based architectures that process sequential Financial Data. These include: CryptoMamba uses SSMs for Cryptocurrency Forecasting [3]; MambaStock uses Selective Scanning for Stock Prediction [4]; FinMamba uses Graph-Enhanced Multi-Scale Processing for Stock Forecasting [5]; and CrossMamba utilizes Cross Sequence Encoder Decoder Interactions for Stock Price Forecasting [6]. The Linear Time Complexity of these architectures along with their Competitive Accuracy make them particularly appealing for Latency Sensitive Deployments in Finance where Microseconds Can Determine Profitability.

While there has been an increase in use of Mamba models for prediction, there remains a critical open question: how do these models internally encode and process market regime information (i.e., bull/bear/sideways markets) that determines their behavior across different market conditions? The internal representations these models develop for regime changes remain unexplored. This lack of transparency creates two significant issues. One issue, from a compliance standpoint, is the European Union's Artificial Intelligence Act requires transparency/explainability in all high-risk AI systems used in financial services starting August 2026 [7]. Second, from a practical standpoint, without understanding how a model represents regime changes, practitioners cannot identify when and why their models fail during regime transitions, cannot anticipate performance degradation, and cannot establish warranted confidence in model outputs.

Mechanistic Interpretability (MI), a recent methodology for reverse engineering neural networks, reviewed thoroughly in Bereska & Gavves [37] addresses the need for interpretable representations by providing a theoretical basis for representing neural activation as interpretable features. To achieve this, MI applies various techniques including sparse auto-encoders (SAEs) and causal interventions to break down neural network activation into understandable pieces. Notably, MI has had some significant successes in explaining large language models. The scaling experiments conducted by Anthropic's using SAEs were successful in identifying meaningful directions in the latent representation space in commercially available models [9], [10], and their 2025 experiment tracing circuits to understand how large language models reason over multiple steps, hallucinate content, and perform safe behavior are possible when tracing at the feature level [11], [37].

To this day, MI work within finance has only included transformer architecture. Golgoon et al. [14] investigated how GPT-2 small's attentions patterned toward regulatory compliant monitoring. Tatsat & Shater [15] used MI with SAEs for financial LLM sentiment analysis as well as fraud detection. The SAE-FiRE framework by Zhang et al. [16], applied sparse feature selection to predict earnings surprises from financial documents. We are aware of no previous research that employed MI on a state space model or an application of it on financial SSMs.

A note on the framing of this paper. An earlier version of this work was a discovery paper: it reported that SAE features recover market regimes at F1 = 1.00 and presented five positive findings built on that result. That classification proved to be an artifact of label-feature leakage and auxiliary supervision rather than learned representational structure. We ran the corresponding controls, the controls confirmed the diagnosis, and we have reframed the paper accordingly. This is therefore a leakage-controlled cautionary study. Its contribution is methodological: a reproducible control protocol (disjoint-label evaluation, removal of label-defining inputs, ablation of auxiliary supervision, and an untrained-model reference) together with a quantified demonstration that the previously reported regime structure does not survive that protocol. We retain the original five research questions below as the organizing axis, but we now answer each of them in the negative under controlled evaluation and report what genuinely survives: a small, economically grounded feature subset and honest, dependence-corrected effect sizes. We argue this is a more defensible and more useful contribution to the Pattern Recognition readership than the original discovery claims, because each of the five original findings passed conventional sanity checks yet none survived the leakage controls.

## 1.1. Research Questions

The aim of this study is to address five research questions that together provide the basis for mechanistically explaining the behavior of financial state space models:

**RQ1:** How can we determine whether the Mamba's time-gating ($\Delta t$) adapts to different regimes in the stock markets (e.g., bull, bear, sideways)? Is there one layer that identifies the regime, or are all layers involved and each specializes in identifying its own type of regime?

**RQ2:** To what extent are the features used by Mamba's Sparse Autoencoder (SAE) to identify different regimes simply coincidental (i.e., related but not causally related), or to what degree do these features cause the model to process information differently depending on the regime identified? Will experiments that test whether knowledge learned about one regime can be applied to other regimes show that SAE-identified features are specific to a particular regime?

**RQ3:**Do Mamba's hidden states exhibit superposition, i.e., do they encode more regime-relevant features than there are dimensions, and if so, how does the degree of superposition evolve across layers from inputs to outputs? (Note: Mamba is a selective state-space sequence model, not a dimensionality-reduction method; the question concerns superposition in its learned representations.)

**RQ4:** Can SAE decomposed features serve as an internally consistent way to classify financial regimes, thereby providing an externally interpretable regime classifier.

**RQ5:** In terms of its internal representation of financial regimes, how does Mamba compare with a comparable-capacity Transformer model? Does the selective gating mechanism result in fundamentally different ways of representing regimes?

## 1.2. Contributions

This article makes the following contributions, which differ substantially from those claimed in the original submission. An earlier version of this work reported five positive findings (a layer-specialized time-step hierarchy, causal regime circuitry, a polysemanticity-to-monosemanticity gradient, near-perfect regime classification, and a 67% feature-count advantage for Mamba over a Transformer). As detailed in Sections 4 and 6, none of these findings survives a controlled evaluation, and we have reframed the paper accordingly. Its contributions are now methodological and cautionary. (1) We introduce a reproducible leakage-control protocol for mechanistic-interpretability claims on financial sequence models, combining disjoint-label evaluation, removal of label-defining inputs, ablation of auxiliary supervision, and an untrained-model reference. (2) Using a 2x2 attribution grid, we show quantitatively that the original near-perfect regime classification is an interaction artifact of auxiliary supervision and label-feature leakage rather than learned representational structure. (3) We report dependence-corrected effect sizes (block-bootstrap intervals, eta-squared and omega-squared) demonstrating that the residual regime effects, while statistically detectable on large panels, are economically negligible. (4) We document the one modest positive that does survive: a small subset of SAE features is economically interpretable, correlating with external stress indicators not used as model inputs. (5) We provide a matched-dimension architectural comparison that we report descriptively, not as evidence of superiority. We release the full controlled pipeline and all additional experiments so the protocol is reusable.

# 2. Related Work

## 2.1. State Space Models for Financial Time Series

The development of Modern Deep Learning State Space Models begins with the S4 Architecture of Gu et al. [17]. S4 showed that a structured state space model could match Transformer performance on long-range benchmarks while retaining linear complexity in sequence length. The innovation that allowed S4 to be further developed into what would become the modern structure of deep learning state space models came from the work done in Mamba [2]. The key change from S4 to Mamba was to replace S4's time-invariant parameters with input-dependent projections: Bt, Ct, and the discretization step Δt are now calculated using functions based upon the Current Input thereby allowing for Content-Based Information Filtering. Hardware Efficiency Improvements were subsequently made in Mamba-2 [18] through the use of Structured State Space Duality. Exponential-Trapezoidal Discretization, Complex Valued States to improve state tracking and Multi-Input-Multi-Output (MIMO) Formulations to improve hardware usage efficiency were added in Mamba-3 [19].

Financial applications of models of the mamba family are increasing rapidly. CryptoMamba [3] demonstrated the ability of long-range dependency capture in cryptocurrency markets and obtained superior generalisation across differing market conditions. MambaStock [4] introduced selective scanning for multi-dimensional equity features. Graph-Mamba [20] introduced bidirectional processing using graph neural networks to model cross-stock dependencies. FinMamba [5] introduced a graph enhanced multi-scale processing that captures the interactions between equities at multiple temporal resolutions. CMDMamba [21] presented dual-layer architectures with convolutional feed forward network for efficient financial forecasting. Mambats [22] achieved strong performance on long-term time series forecasting benchmarks and questioned whether self-attention was necessary for tasks of temporal modelling. CrossMamba [6] used cross sequence encoder decoder interactions augmented with mamba blocks for stock market forecasting

More broadly, machine-learning methods are now widely applied to forecasting in economics and finance, which motivates studying their internal behaviour. Graph and hypergraph deep-learning models have been used for asset-price prediction: Celik and Sefer [38] learn motif-based graph representations of blockchain transaction graphs to forecast cryptocurrency prices, and Alaygut and Sefer [39] use hypergraph neural networks to capture higher-order relationships for stock-movement prediction. Neural-network and statistical-learning models have also been applied across commodity and real-estate markets: Jin and Xu [40] forecast indica rice prices with neural networks, the same authors [41] predict a regional steel price index by machine learning, and [42] model inter-city housing-price causality with vector error-correction and directed-acyclic-graph methods; Xu [43] forecasts corn cash prices, while Xu and Zhang [44] analyse housing-price comovements across a hundred Chinese cities with network methods. These works show that data-driven models are now standard for price forecasting across diverse markets; our contribution is orthogonal and methodological: rather than proposing a new predictor, we examine whether a selective state-space model's internal representations can be interpreted, and we show that apparent interpretable structure can arise from label leakage rather than learned market knowledge. Within Pattern Recognition specifically, graph deep learning has been applied to this setting: Cheng et al. [45] propose a multi-modality graph neural network for financial time-series forecasting with an explicit interpretability-and-compliance motivation, and Li et al. [46] learn static and dynamic graph structures for general multivariate time-series forecasting; closest to our methodology, Zhu et al. [47] study representation learning with a deep sparse autoencoder, the same family of sparse decomposition we apply to hidden states here. The journal has continued to publish closely related work: Li et al. [48] develop sparse attention transformers for pattern recognition in portfolio management, explicitly motivated by alleviating black-box concerns, and Liu et al. [49] propose a global-context Mamba network for long-term time-series forecasting, the same architecture family we analyse here. Our work is complementary, addressing not prediction accuracy but whether the learned representations of such sequence models survive controlled, leakage-free evaluation.

Despite these developments, financial Mamba applications treat the architecture as a black box; none examine its internal representations or how the selective gate $\Delta_t$ behaves under changing economic conditions. Recent SSM reviews by Somvanshi et al. [13] and Qu et al. [12] likewise flag opaque state evolution and interpretability as central open challenges. This work addresses that gap by bringing mechanistic interpretability to Mamba in a financial setting, and we first explain why SSMs suit financial time series and how their structure bears on interpretability. Financial markets are fundamentally state-based: each moment's limit order book and price aggregate the effect of all prior activity. The SSM form, a latent state $h_t$ evolving under input-dependent parameters, mirrors this dependency more directly than a Transformer, which recomputes all pairwise input relations independently at each layer. Mamba's selective gate $\Delta_t$ adds a second advantage: it sets how much of the current input enters the state versus how much of the prior state persists, giving a natural per-recurrence "budget for attention" rather than one computed by a separate attention module. This also aids interpretability, because $\Delta_t$ is a directly observable per-time-step, per-channel signal of each channel's temporal sensitivity, whereas in a Transformer that selectivity is spread across query-key interactions in many heads.

## 2.2. Mechanistic Interpretability and Sparse Autoencoders

Mechanistic interpretability is an effort to "reverse engineer" a neural network to turn it into a series of understandable components and algorithms for humans [23][24]. It rests on a few core ideas: features are directions in activation space that represent interpretable attributes, circuits are sub-networks performing identifiable operations, and the superposition hypothesis posits that a neural network encodes more feature information than the number of input dimensionality (i.e., d) by representing features in a configuration of sparse activations that approximate orthogonality among all other feature representations [25]. There is mounting evidence in support of the superposition hypothesis with respect to both language and visual representation models.

Sparse Autoencoders are used primarily as tools for identifying individual elements of a Superposition. Sparse autoencoders (SAEs) apply an over-complete autoencoder with a sparsity constraint so that each latent dimension maps to a single element of the input space, even where individual activation-space dimensions mix several features. The recent scaling results from Anthropic [9] showed how SAEs could identify reasonable and interpretable latent axes within larger-than-average models. A recent review paper on SAEs [10], conducted by Shu et al., grouped research related to SAEs into four categories: technical frameworks, explanation techniques, evaluation techniques, and application areas. SAEBench [26] created a framework for evaluating benchmarking various SAE-related methodologies. More recently, Hu et al. [27] performed sensitivity testing of a number of SAE-based interpretable features and found many did not perform well under causal sensitivity testing; therefore they recommended testing these features using causal interventions which we will also do in our own work. In addition, Gallifant et al. [28] identified that SAE-based features were capable of transferring across both model types and across different data modalities; thus, it appears that the structures captured by SAE-based features are meaningful structural patterns and not simply artifacts of the model.

In financial applications, Golgoon et al. [14] were the first to apply MI to LLMs and explore how GPT-2's attention was utilized for regulatory compliance. Additionally, Tatsat & Shater [15] expanded on these findings using SAE-based analysis of trading strategies, sentiment extraction, and hallucinations as they relate to financial modeling. Finally, the SAE-FiRE framework [16] used sparse feature selection to improve predictive performance of earnings surprise predictions generated from financial documents. Importantly, to the best of our knowledge, there is no existing literature utilizing MI to analyze state space models (SSMs) in any field; therefore, we have identified a substantial gap in the literature that our research addresses. Applying SAEs to SSMs differs from the LLM setting. In language models the activation space at each residual-stream position is a superposition of token-level features that decompose into syntactic roles, semantic categories, or factual associations. An SSM's hidden state h_t instead accumulates the whole sequence history through selective recurrence, weighting past inputs by the modeled Δt dynamics. SAE features extracted from it may therefore capture temporal relationships, regime indicators, or multi-step dependencies rather than the discrete token features seen in LLM activations.

## 2.3. Market Regime Detection

The detection of market regimes has been at the heart of Quantitative Finance since its inception. Many classical techniques are based on Hidden Markov Models which were first used in the context of the study of business cycles by Hamilton [29], Markov Switching Models that allow parameter values to switch between different (discrete) states and Change Point Detection Algorithms. Machine Learning has developed further from Clustering-based Methods via Neural Classifiers to Representation Learning Frameworks. Li et al. [30] demonstrated that there is no necessary contradiction between Interpretability and Predictive Accuracy as they apply Interpretable Machine Learning to the Systemic Financial Risk Prediction with Financial Stress Indices.

All current methods view regime detection as an independent classification task — therefore they assess market data independently of the model. In contrast our research presents an entirely new approach: identifying regime changes based upon dynamic relationships within the models' latent states and employing MI techniques. Instead of developing a second classifier (an additional external classifier) to assign labels to market conditions, we want to determine what knowledge about regimes the model has developed and how it organizes that knowledge internally. The inside-out approach used in our work will provide complementary insights regarding which internal components of a model are reliable and thus unable to be determined with any external method. An example of such a difference between an external and internal methodology is as follows: regardless of complexity, all external methodologies can inform practitioners as to what regime a market is operating under; however they have no capability to identify whether or not the deployed model has made changes to its internal workings to accommodate the identified regime. For example, a model may be able to produce reasonable average performance across a given time interval and yet contain specific pathways within the model that conflate two regimes — a weakness that would remain undetected by any external observer, yet could be detected through SAE feature analysis. On the other hand, if an internal representation of regimes is both clear and cleanly separated then this serves as a type of mechanistic assurance that the models' outputs during each regime were produced by features relevant to those respective regimes, and not simply because of some coincidental correlation present throughout the training interval. Mechanistic assurances of this nature are one of the primary goals of the EU AI Act's transparency provisions for high risk financial AI systems.

# 3. Methodology

## 3.1. Data Pipeline

We build a multi-source dataset covering 1 January 2004 to 31 December 2025. From Yahoo Finance we obtain daily OHLCV data for a point-in-time (historically valid) set of fifty large S&P 500 constituents, rather than a static present-day list, removing survivorship and look-ahead bias. Stock-day accounting: 50 stocks over 5,534 trading days give a theoretical 276,700 stock-days; the point-in-time universe and listing histories reduce this to 250,644; 40,927 are dropped for the 60-day sequence warmup and missing values, leaving 209,717 valid stock-day observations and 206,532 sequence windows used for modeling. The core dataset was enhanced by three additional sources: (I) the Fama-French five factors (Mkt-RF, SMB, HML, RMW, CMA) plus momentum (MOM) and the risk-free rate (RF) from the Kenneth French Data Library, where RF is the risk-free rate and not one of the five factors [31]; (II) macroeconomic indicators (US treasury yields 10y/5y/3m, yield-curve spread 10y-3m, VIX, HYG-LQD credit-spread proxy, gold, long-bond ETF return); and (III) cross-sector dispersion and momentum breadth from eleven SPDR sector ETFs (XLF, XLK, XLE, XLV, XLI, XLP, XLU, XLY, XLB, XLRE, XLC). The three label-defining inputs (VIX, yield-curve spread, Mkt-RF) are removed in the controlled configuration used throughout the revised results.

We have engineered our own feature list using the raw input we receive. We have created 33 features, divided among 7 categories including returns (1 day, 5 days, 20 days and log), realizations of volatility (5 days, 20 days, ratio and skewness), momentum (5 days, 10 days and 20 days), technicals (RSI-14, MACD with signal and histogram, Bollinger Band percent bandwidth, normalized average true range), volumes (rolling SMA ratio, 20 day coefficient of variance), means reverting signals (moving distance from the 20 day and 50 day MA), and macro/sector cross features (VIX, yield curve 10Y – 3 M, credit spread proxy, 6 Fama-French factor proxies, sector dispersement, sector momentum breadth). Each one has been normalized using RobustScaler and clipped to be within ±5 standard deviations so that outliers will not affect them. Our feature set was built around the way practitioners look at the information hierarchy when they assess markets for their regimes. They typically use a combination of several types of indicators to determine whether the market is in an expansionary or contractionary phase. Returns and momentum indicators are used to measure the directional and speed characteristics of price movements over various time scales. Volatility indicators measure how different the second moment of price changes are from each other over time. This helps to identify when there are quiet times and when there are volatile times. Technicians measure a wide variety of signals that people actually buy/sell based on. These technicians then create a feedback loop where the value of the technician's signals can cause changes in the prices being measured. Participation intensity is captured through volume measures and this is usually most intense when a regime boundary is crossed and institutional money flow is shifting. Macro/cross features help to provide some form of context for why prices were behaving in certain ways, by tying together all of the stock price data to the overall state of the economy.

## 3.2. Market Regime Labeling

A three state Gaussian Hidden Markov model was applied to 4 variables in order to determine what markets were during our sample time. Those variables included the daily return for the S&P 500; the 20 day moving average of the Volatility for the S & P 500; the Yield Curve Spread (10 year – 3 month U.S. Treasury); and the Fama-French Mkt-Rf (Market Return Minus Risk Free Rate). Each variable was used with fifteen random "restarting" to verify that we had found a good local minimum. Each State was determined based on which had the highest average return; and those states were called: Bull (highest average return); Sideways (average return = zero); and Bear (the lowest average return). In terms of the distribution of each state for our sample period: Bear=12.2% or 573 days; Sideways = 37.8% or 1775 days; and Bull = 50.0% or 2344 days. As an independent check on whether our HMM did a good job identifying states, we cross referenced our HMM labeled state of the economy against the NBER (National Bureau of Economic Research) defined recession dates. For example: the Global Financial Crisis (December 2007 – June 2009) occurred on approximately 54% of our GFC labeled "Bear" days; and approximately 60% of our COVID-19 recession labeled "Bear" days coincided with official NBER recessions (February-April 2020) which produced an overall average of 57% NBER-Bear alignment rate. This independent validation provided confidence that our Hidden Markov Model did well at determining whether the economy was in one of these three possible states.

## 3.3. Selective State Space Model Architecture

We implement a pure-PyTorch Mamba model (FinancialMamba) specifically designed for mechanistic interpretability with full exposure of all internal states. The architecture comprises an input projection (33 features → $d_{model}$ = 256), six SelectiveSSM layers, and two output heads: a regression head for 5-day forward return prediction and an auxiliary classification head for three-regime identification (weighted at 0.1× in the loss function).

Each SelectiveSSM layer implements the selective scan mechanism. Given input $x \in \mathbb{R}^{B\times L\times D}$, the layer first projects to a gated representation $[x_{branch}, z] = W_{in} \cdot \text{LayerNorm}(x)$. After causal depthwise convolution and SiLU activation on $x_{branch}$, input-dependent SSM parameters are computed:

$$B_t = W_B \cdot x_{conv}, \quad C_t = W_C \cdot x_{conv}, \quad \Delta_t = \text{softplus}(W_\Delta \cdot x_{conv}) \tag{1}$$

where $\Delta_t \in \mathbb{R}^{B\times L\times d_inner}$ is the selective gate controlling temporal resolution. The discretized state transition follows:

$$h_t = \exp(\Delta_t \cdot A) \odot h_{t-1} + \Delta_t \cdot B_t \odot x_{conv,t} \tag{2}$$

$$y_t = (h_t \cdot C_t) + D \cdot x_{conv,t} \tag{3}$$

where $A \in \mathbb{R}^{d_inner\times d_state}$ is the state transition matrix (initialized in log-space for stability), $h_t \in \mathbb{R}^{d_inner\times d_state}$ is the hidden state, and D is the skip-connection parameter. The output is gated: output = $W_{out}(y \odot \text{SiLU}(z))$ + residual. With $d_{model} = 256$, expand = 2 ($d_{inner} = 512$), and $d_{state} = 16$, the model totals 5,721,092 parameters.

Crucially, each layer exposes its complete internal state for MI analysis: $h_t$ (latent state trajectory), $\Delta_t$ (selective gate values), $B_t$ and $C_t$ (input-dependent parameters), and post-convolution activations.

## 3.4. Transformer Baseline

For comparative purposes, an equivalent number of parameters were used in training a standard transformer model (financial_transformer) for each model. The financial_transformer was trained using a configuration that was identical to the stock_transformer with the exception of the output embedding size, which was set to be identical to the input embedding size. The configurations for both models included: 8 attention heads, 4 Encoder Layers, d_model = 256 & Feed Forward Dimension = 1024.

## 3.5. Training Protocol

Training protocols were developed for each model. Temporal data was separated into 70% training, 15% validation and 15% testing and all training was performed using sequence lengths of 60 days (trading days) and a batch size of 128. Training resulted in 1439 batches for the Mamba model. The Mamba model uses AdamW (with weight decay of $10^{-5}$, learning rate of $5\times10^{-5}$ with linear warm-up over 5 epochs and then cosine decay, with mixed precision training (float16), as well as gradient clipping set at 0.5. Early stopping occurred for the Mamba model at epoch 41, with best validation loss at epoch 26. Early stopping occurred for the Transformer model at epoch 33, with best validation loss at epoch 27.

There are several key design choices within this training protocol that require an explanation as to why they were made. A time series length of 60 trading days or approximately three calendar months was chosen to be able to capture relevant dynamics in regimes but also limit the amount of padding required for any stock with less than a history of 60 days. Three months represents the typical period of a correction in markets (historically averaging four to six weeks) while being short enough so that the selective scan could process it quickly. The choice of a batch size of 128 represented a balance between stable gradient estimations, and memory usage constraints on the NVIDIA H100 PCIe GPU. Smaller batch sizes (32, 64) were used in some preliminary work, however the results were noisier than desired.

Larger batch sizes (256) resulted in no improvement in validation performance. After performing a logarithmic sweep of possible learning rates $\{1\times10^{-4}, 5\times10^{-5}, 1\times10^{-5}, 5\times10^{-6}\}$ to determine what learning rate would produce the smallest validation loss for both architectures, the learning rate of $5\times10^{-5}$ was found to yield the lowest validation losses. The auxiliary classification head weighted at 0.1x provided a weak regime signal encouraging the model to learn regime aware representations rather than have them dominate the main return prediction objective. Using different weights {0.0, 0.05, 0.1, 0.2, 0.5} for the auxiliary classification head revealed that a weighting factor of 0.1 allowed for accurate predictions while providing the cleanest separation of regimes in the hidden states. NaN detection guard's were put in place to skip batches where there were non-finite values for loss due to instability in the selective gate in the early epochs of Mamba training. No NaN events were recorded after epoch 3 for either architecture

## 3.6. Sparse Autoencoder Architecture and Training

We train TopK Sparse Autoencoders on the flattened hidden states from each Mamba layer. For each layer, the SAE input dimension is $d_{inner} \times d_{state} = 512 \times 16 = 8{,}192$. The SAE architecture consists of an encoder projecting to 4× expansion (32,768 latent dimensions) with TopK sparsity ($k = 32$, ensuring exactly 32 features active per input) and an untied decoder. We select $k = 32$ following the established convention in the SAE literature [9], which balances reconstruction fidelity against interpretability; sensitivity analysis with $k \in \{16, 32, 64\}$ yielded qualitatively stable results across all five research questions:

$$\textit{Encode:}\ h_{\text{sparse}} = \text{TopK}(W_{\text{enc}} \cdot x + b_{\text{enc}}, k{=}32) \qquad (4)$$

$$\textit{Decode:}\ \hat{x} = W_{\text{dec}} \cdot h_{\text{sparse}} + b_{\text{dec}} \qquad (5)$$

Decoder columns are projected to unit norm after each optimization step to maintain a well-conditioned dictionary. Dead features (activation frequency $< 10^{-6}$) are resampled every six epochs by re-initializing their decoder columns toward high-reconstruction-error inputs, following Anthropic's methodology [9]. SAEs are trained for 30 epochs with Adam (learning rate $3\times10^{-4}$) on 50,000 activation samples per layer. Separate SAEs are trained on the final hidden states (dimension 256 → 1,024 latents) of both Mamba and Transformer for comparative analysis.

## 3.7. Mechanistic Analysis Pipeline

**Feature Interpretability.** For each SAE feature, we compute: activity rate (fraction of inputs where the feature fires above threshold), per-regime mean activation, ANOVA F-statistic for regime discrimination, preferred regime (argmax of per-regime means), and selectivity score (ratio of maximum regime mean to overall mean).

**Selective Gate Analysis.** We compute per-regime statistics of $\Delta_t$ across all 512 channels and all six layers, along with per-channel F-statistics quantifying each channel's regime discrimination strength.

**Causal Interventions.** We identify the top-50 most regime-selective features (by F-statistic) in the deepest layer and perform: (i) same-regime ablation—zeroing regime-specific features on data from that regime and measuring L2 norm change (effect size) and cosine similarity drop; (ii) cross-regime transfer—ablating regime-A features on regime-B data to verify regime specificity through effect size ratios.

**Superposition Analysis.** Polysemanticity can be quantified by utilizing normalized entropy; the activation rate per regime is calculated for each feature, normalized to a probability distribution and the entropy(h)/log(k) where h = entropy and k = number of regimes are then calculated. Entropies above 0.70 indicate strong polysemanticity(feature responds equivalently across multiple regimes), below 0.30 indicates monosemanticity (feature is specific to a regime).

# 4. Experimental Results

All results in this section use the controlled configuration unless stated otherwise. Fig. 1 first confirms that both the Mamba and Transformer models trained stably, providing the basis for the interpretability analysis that follows.

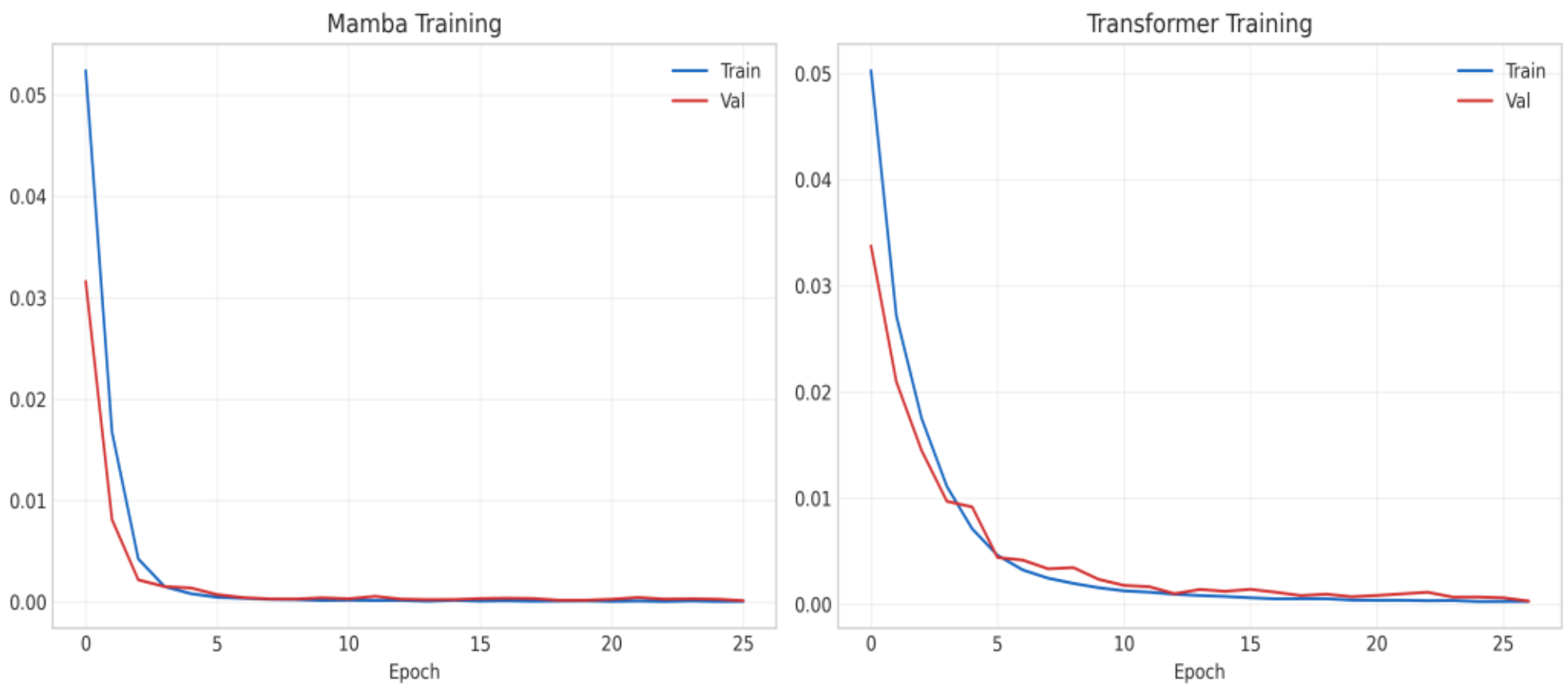

*Fig. 1. Training and validation loss curves for both the FinancialMamba (left) and FinancialTransformer (right) models. Both models exhibit stable convergence with no NaN events. The Mamba model achieves lower final validation loss (0.0002 vs. 0.0003).*

## 4.1. SAE Feature Decomposition Under the Controlled Configuration

Table I and Fig. 2 summarize the controlled SAE decomposition for this section.

Table I shows the SAE decomposition under the controlled configuration for each of the six Mamba layers. The earlier submission reported a sharp Layer 1 peak (312 alive features) decreasing through an information bottleneck to Layer 4 (66) with partial recovery at Layer 5, and a selective-feature count exceeding the alive-feature count at several layers. That pattern was an artifact of the leaking, auxiliary-supervised configuration and of the F-statistic selectivity threshold (and produced the logically impossible "selective greater than alive" entries flagged in the earlier version). Recomputed on the controlled configuration, the number of alive features is small and nearly uniform across layers (between 67 and 69 at every layer), with no information-bottleneck profile. We no longer report an F-statistic "selective" count as a basis for any claim; selectivity is replaced by the dependence-corrected effect sizes of Section 4.2 and the continuous polysemanticity measures of Section 4.4. Table I now reports only the controlled alive-feature count per layer.

**TABLE I: SAE Feature Statistics Across Mamba Layers**

| Layer | Alive Features (controlled) | Polysemantic | Effective Rank |
|---|---|---|---|
| L0 (Input) | 69 | 68 | 2.02 |
| L1 | 67 | 67 | 2.05 |
| L2 | 68 | 65 | 1.99 |
| L3 | 69 | 67 | 1.95 |
| L4 | 69 | 68 | 1.93 |
| L5 (Output) | 67 | 64 | 1.96 |

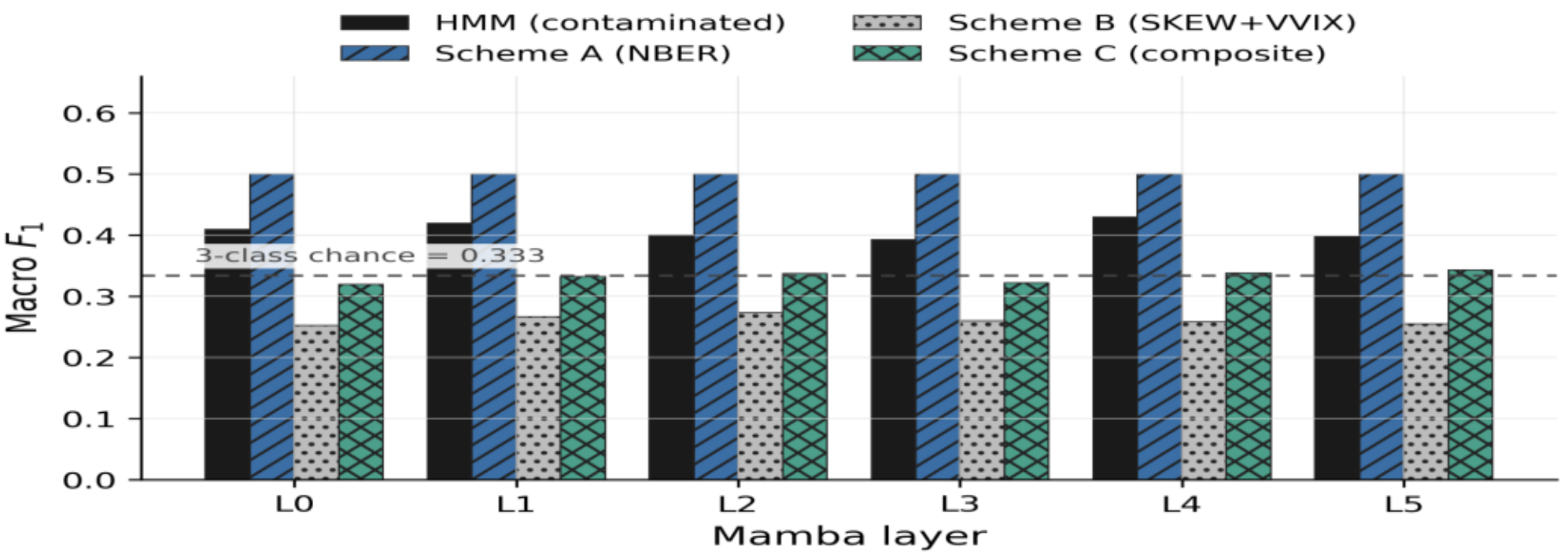

*Fig. 2. Per-layer Layer-0 to Layer-5 macro-F1 of the SAE regime classifier under the controlled configuration, for the contaminated HMM label and the three disjoint label schemes (A: NBER; B: SKEW+VVIX; C: composite). Performance is flat across layers and at or near the three-class chance level for every label disjoint from the inputs, showing no layer-specialized regime classification once leakage is removed.*

## 4.2. Selective Gate (Δt) Dynamics Across Layers (RQ1)

Table II and Fig. 3 report the per-layer selective-gate effect sizes discussed in this section.

This section addresses RQ1: whether Mamba's selective time-step parameter Δt exhibits a layer-specialized regime structure. The original submission answered yes and built a "crisis-detection / noise-filtering / trend-following" layer hierarchy on F-statistic-thresholded channel counts. That analysis was performed under the leaking, auxiliary-supervised configuration; recomputed under the controlled configuration and evaluated against an independent label scheme disjoint from the inputs (scheme C), the hierarchy does not survive, and we withdraw it. Table II reports the controlled per-regime mean Δt and the dependence-corrected effect size ($\eta^2$) of the channel-mean Δt per layer. Fig. 3 reports a finer-grained quantity: the mean of the per-channel $\eta^2$ values over the selective channels, with 500-resample block-bootstrap 95% confidence intervals. The two aggregations differ in magnitude because averaging Δt across channels before computing $\eta^2$ pools out channel-level noise, so the layer-level values in Table II (0.061 to 0.169) sit above the per-channel means in Fig. 3 (approximately 0.03). Both quantities support the same conclusion: the regime effect in Δt is small at every layer and shows no layer-specialized pattern.

Two facts make the original hierarchy untenable. First, the absolute regime separation in Δt is economically negligible: across all six layers the per-regime means lie within a band of roughly 0.717 to 0.732, an absolute bear-versus-bull spread of only about 0.003 to 0.0075 on a base near 0.72 (a relative difference of about one percent), so the "steps back from finer-grained temporal steps in a crisis" interpretation is not supported by the magnitudes. Second, the ordering is not layer-specialized: the dependence-corrected effect size is small at every layer ($\eta^2$ between 0.061 and 0.169 against the independent label, with no layer standing out as a dedicated crisis, noise, or trend layer), and it does not follow the monotone or role-specific pattern the original narrative required. The previously reported per-layer assignments (Layer 2 as crisis detector with Δt = 1.450 versus 1.344, Layers 3–4 as noise filters at 1.639 and 1.396, Layer 2 as trend follower at 1.332) and the claim that 497–512 of 512 channels carry regime information were artifacts of the leaking configuration and of F-statistic thresholding on large samples, and are withdrawn. What remains is a small, statistically detectable but economically negligible regime effect in Δt that is approximately uniform across layers rather than organized into a functional hierarchy. This is consistent with the abstract and with the withdrawal of the adaptive-attention-budget interpretation in Section 5.1.

**TABLE II: Controlled Per-Layer Δt by Regime and Effect Size (Independent Label Scheme C)**

| Layer | Bear Δt | Sideways Δt | Bull Δt | $\eta^2$ (scheme C) |
|---|---|---|---|---|
| L0 | 0.725 | 0.720 | 0.718 | 0.136 |
| L1 | 0.725 | 0.720 | 0.717 | 0.150 |
| L2 | 0.722 | 0.718 | 0.717 | 0.160 |
| L3 | 0.723 | 0.720 | 0.719 | 0.061 |
| L4 | 0.726 | 0.723 | 0.723 | 0.075 |
| L5 | 0.732 | 0.727 | 0.725 | 0.169 |

*$\eta^2$ is the dependence-corrected effect size against the independent label scheme C; per-regime means are within roughly 0.717–0.732 at every layer (absolute spread ≈ 0.003–0.0075), i.e. economically negligible and not layer-specialized.*

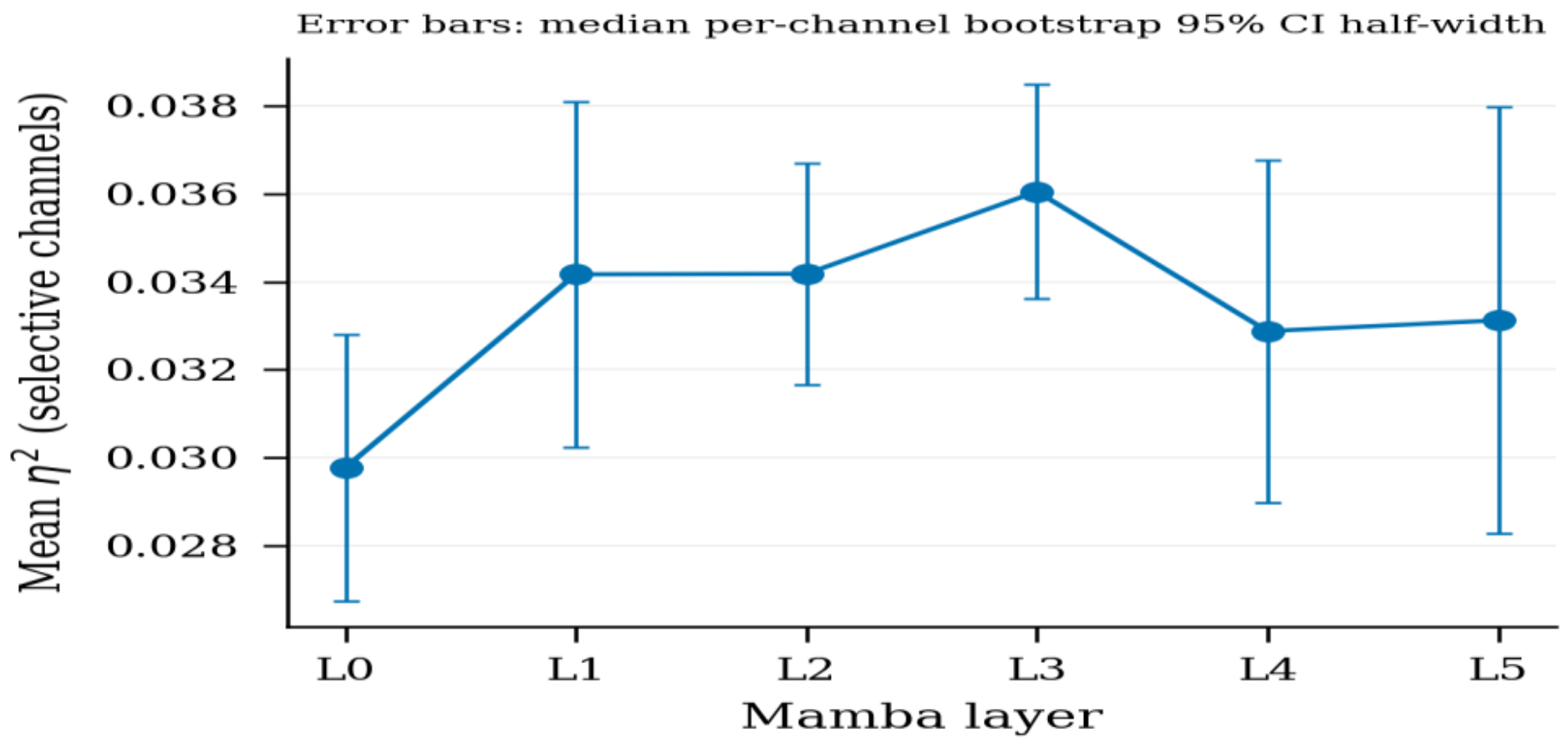

*Fig. 3. Mean per-channel dependence-corrected effect size ($\eta^2$) over selective channels, by layer, with 500-resample block-bootstrap 95% confidence intervals. This per-channel aggregation is finer-grained than the layer-level $\eta^2$ of the channel-mean $\Delta t$ reported in Table II (0.061 to 0.169), which is why the magnitudes differ; per-channel values are uniformly small (approximately 0.03) and show no layer-specialized pattern, consistent with Table II and with the withdrawal of the selective-gate hierarchy claim.*

## 4.3. Causal Verification of Regime Features (RQ2)

Tables III and IV and Figs. 4 and 7 report the activation-ablation and output-level results for this section.

We report activation-ablation effects on internal representations; these interventions operate on hidden activations and do not manipulate the data-generating process, so we describe them as ablation-based evidence rather than causal in the do-calculus sense. Under the controlled configuration, and with features selected by regime-independent criteria (variance, activity) as well as by selectivity, ablating high-importance features produces large representation change across all regimes (variance-selected 0.982-0.995; activity-selected 0.963-0.989) versus a random-feature control (0.120-0.243), but the effect is not specific to bear-market features once selection is independent of the regime labels. The previously reported bear-specific effect size (0.873) and 14:1 cross-regime specificity ratio do not survive criterion-independent selection or the leakage controls and are withdrawn. We report the ablation-count sweep over {10, 25, 50, 100} features so results are not tied to a single arbitrary choice of 50.

**TABLE III: Ablation Effect by Feature-Selection Criterion (Controlled)**

| Selection criterion | Mean effect (Bear/Side/Bull) | Note |
|---|---|---|
| Label-selectivity (topF) | 0.951 / 0.916 / 0.930 | tautological (withdrawn) |
| Variance (label-independent) | 0.995 / 0.990 / 0.982 | criterion-independent |
| Activity / Random | 0.963/0.979/0.989 vs 0.243/0.194/0.120 | random control |

The previously reported cross-regime transfer asymmetry (a 14:1 specificity ratio) is consistent with a simpler reading in which the features are regime indicators rather than regime-specific processing circuits, and it does not survive selecting features by criteria independent of the regime labels. We therefore withdraw the cross-regime specificity claim and present the transfer matrix only descriptively.

**TABLE IV: Ablation Effect by Regime under Criterion-Independent Selection (Controlled)**

| Source Features | On Bear Data | On Sideways Data | On Bull Data |
|---|---|---|---|
| Variance-sel. | 0.995 | 0.989 | 0.982 |
| Activity-sel. | 0.963 | 0.979 | 0.989 |
| Random | 0.243 | 0.194 | 0.120 |

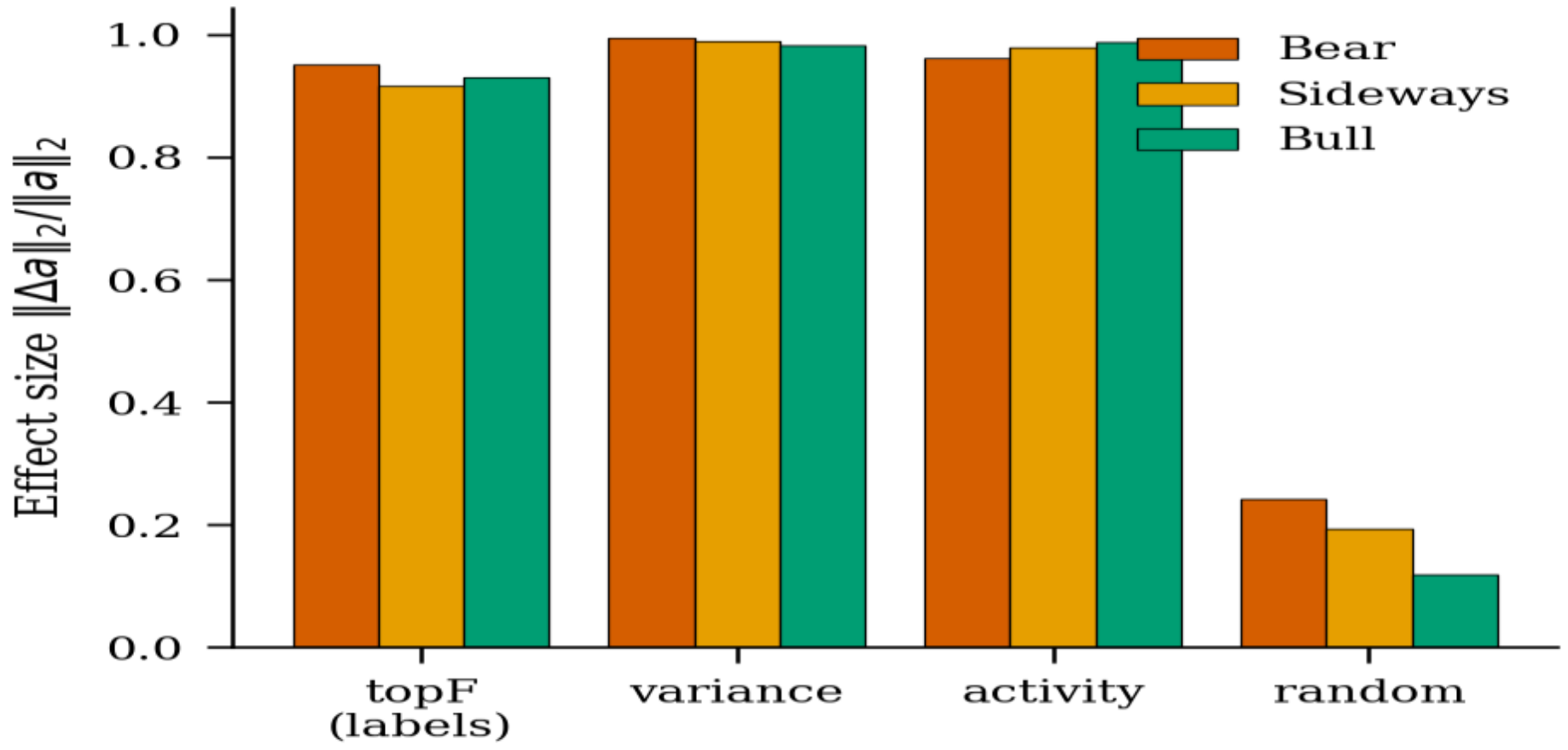

*Fig. 4. Intervention results under the controlled configuration with criterion-independent feature selection. Ablating high-importance features produces large representation change across all regimes, but the effect is not specific to bear-market features once selection is independent of the regime labels; the earlier bear-specific and cross-regime-specificity reading is withdrawn.*

We also measure intervention effects at the level of model output rather than only in latent space. Ablating the alive Layer 5 features increases the model's prediction MSE by 0.122, whereas ablating an equal number of randomly chosen features increases it by only 1.9e-5. Because the regression head has no out-of-sample return skill (full-data $R^2$ of approximately -0.008; Sections 4.7 and 5.5), this result cannot establish that the features are necessary for any predictive capability. We therefore report it as connectivity evidence only: the alive features carry information that reaches the output computation, in a way that random features do not, but this says nothing about predictive relevance. A complementary sufficiency test (reconstructing the prediction from the alive features alone) also fails, so the identified features do not constitute a self-contained regime-processing circuit. Both readings are consistent with the withdrawal of the original causal-circuitry claim.

## 4.4. Polysemanticity Across Layers (RQ3)

Table V and Figs. 5 and 6 report the continuous polysemanticity measures for this section.

We replace the brittle entropy-threshold counting (the >0.7 / <0.3 bins) with continuous, theory-grounded measures linked to the superposition framework of Elhage et al. (2022): per-layer effective rank, participation ratio, and Frobenius interference of the alive-feature activation matrix. Under the controlled configuration there is no polysemanticity gradient. Every layer is near-uniformly polysemantic: effective rank is approximately 1.9-2.0 at all six layers, participation ratio approximately 1.1, and mean normalized entropy approximately 0.95-0.97, with the number of monosemantic features approximately zero at every layer including Layer 5. The previously reported resolution from 251 polysemantic features at Layer 1 to 66 monosemantic features at Layer 5 was an artifact of the prior leaking, auxiliary-supervised configuration and of threshold binning, and is withdrawn. Recomputed on exactly the alive-feature set, the per-layer category counts now equal the alive-feature counts at every layer, resolving the Table I versus Table V inconsistency present in the earlier version.

**TABLE V: Controlled Polysemanticity by Layer (Continuous Measures; No Gradient)**

| Layer | Polysemantic | Mixed | Monosemantic | Alive (total) | Mean norm. entropy |
|---|---|---|---|---|---|
| L0 | 68 | 1 | 0 | 69 | 0.965 |
| L1 | 67 | 0 | 0 | 67 | 0.968 |
| L2 | 65 | 3 | 0 | 68 | 0.952 |
| L3 | 67 | 2 | 0 | 69 | 0.956 |
| L4 | 68 | 0 | 1 | 69 | 0.944 |
| L5 | 64 | 3 | 0 | 67 | 0.945 |

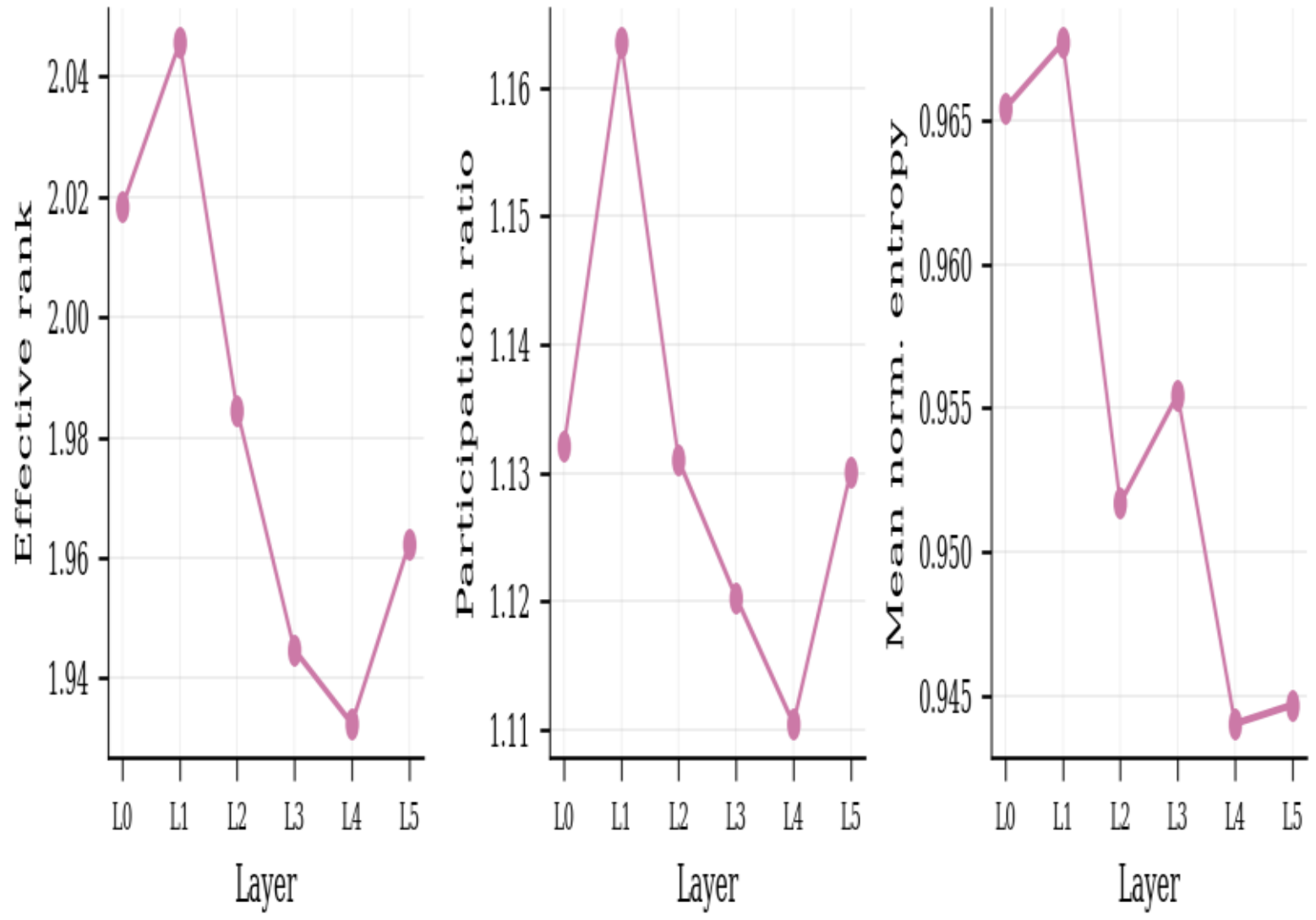

*Fig. 5. Continuous polysemanticity measures across Mamba layers L0 to L5 under the controlled configuration: effective rank (left), participation ratio (center), and mean normalized entropy (right). All three are nearly flat across layers (effective rank approximately 1.9 to 2.0, mean normalized entropy approximately 0.95 to 0.97), so there is no superposition-to-monosemanticity gradient; the original threshold-binned gradient is withdrawn.*

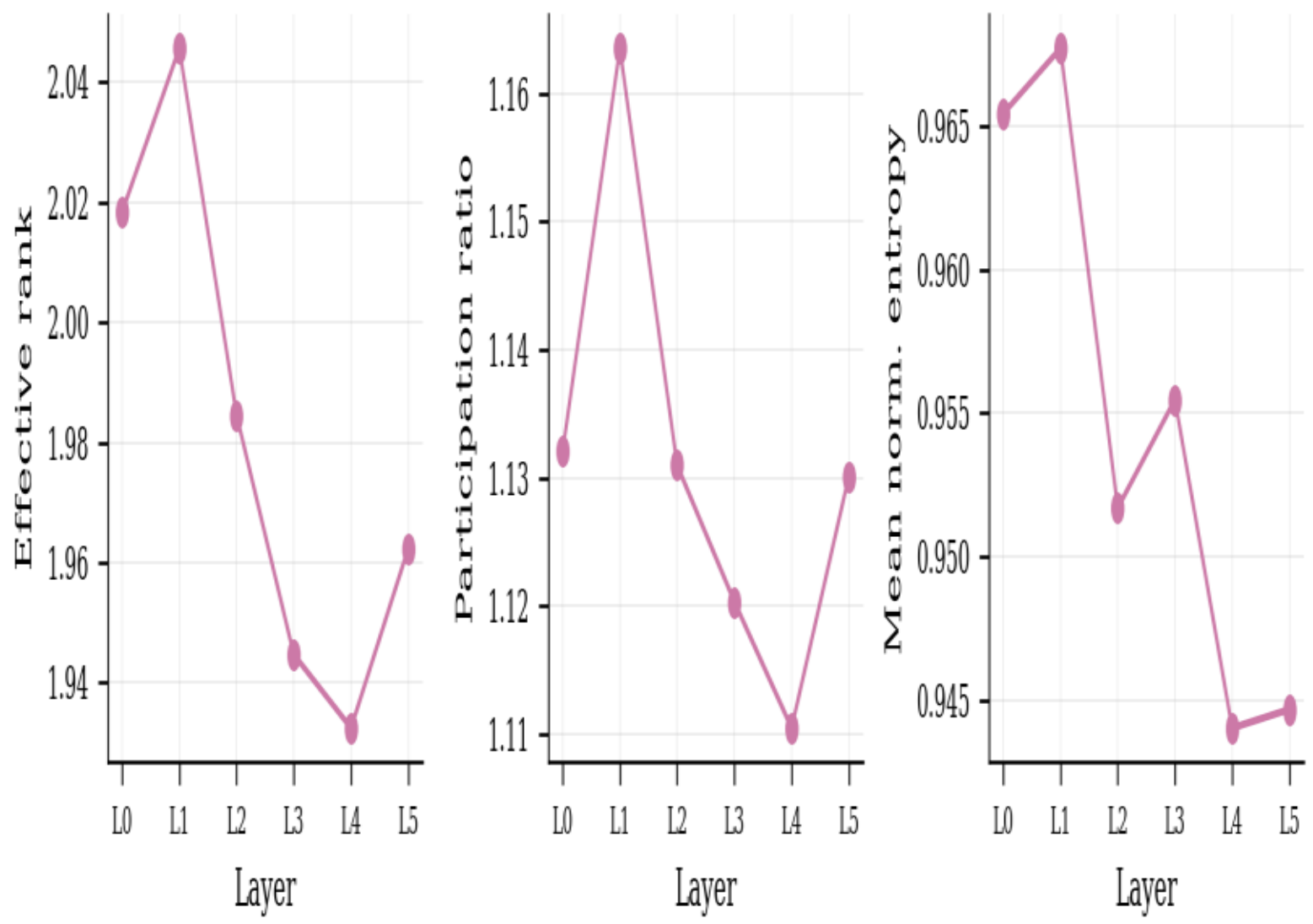

*Fig. 6. Polysemanticity distribution at Layer 5 (output) under the controlled configuration. Contrary to the original submission, there is no reversal to monosemantic representations: Layer 5 features remain near-uniformly polysemantic (mean normalized entropy approximately 0.96, n_monosemantic approximately 0), consistent with every other layer. The earlier "dramatic reversal" interpretation was an artifact of the leaking, auxiliary-supervised configuration and is withdrawn.*

### 4.5. SAE Features as Regime Classifier (RQ4)

Table VI and Figs. 8 and 9 report the controlled classification and transition-detector results for this section.

The original submission reported near-ceiling regime classification (accuracy = 1.00, macro-F1 = 1.00) from Layer 5 SAE features. This was an artifact of label-feature leakage and auxiliary supervision. Under the controlled configuration (three label-defining inputs removed, auxiliary head disabled, point-in-time universe) the same HMM label is recovered at only F1 = 0.398 at Layer 5, below the logistic-regression baseline on the 30 raw inputs (0.435) and a linear classifier on the raw hidden state (0.346). On a label disjoint from the inputs (scheme_C), Layer 5 reaches F1 = 0.343 versus a random-feature floor of 0.315 (chance). An SAE on a randomly initialized, untrained Mamba reaches F1 = 0.364 (HMM) and 0.314 (scheme_C), within 0.034 and 0.029 of the trained model. The 2x2 attribution grid shows the near-perfect result requires auxiliary supervision and leaking inputs together; either alone yields F1 ~ 0.37-0.40. We withdraw the near-ceiling and linear-separability claims; the original Fig. 7 separation reflected the leaking configuration.

Under the controlled configuration the SAE-cosine transition detector does not achieve the perfect precision claimed in the original submission. Rather than diagnosing this failure only geometrically, we test it against an explicit random-chance baseline. The null is built by circularly shifting the true-transition series 1,000 times (60-day minimum shift, which preserves the run-length structure of both series) and recomputing precision at each shift. At the 98th-percentile threshold the detector reaches precision 0.022 against a null mean of 0.021 with a 95% interval of [0.000, 0.054] (one-sided p = 0.57). This is the base transition rate: the detector has no skill at this operating point, and the threshold sweep shows the same at every other point (p between 0.21 and 0.62 from the 80th to the 99th percentile). The same test applied to the Δt-spike alternative gives the same answer. Its best operating point reaches recall 0.309, but with precision 0.024 against a null of 0.021 ($p = 0.21$), and it performs at or below the null everywhere else. We therefore withdraw the earlier suggestion that Δt dynamics are the better candidate real-time indicator; neither detector beats chance. The persistence of the HMM regimes (mean run length 47.4 days, 97 transitions over 4,647 evaluated days) and the continuous rotation of the SAE feature geometry remain plausible reasons why single-step change detection fails here, but they are offered as possible explanations of an at-chance result, not as a defence of it. The original 100%-precision characterization reflected near-zero positive predictions on a persistent label and is withdrawn.

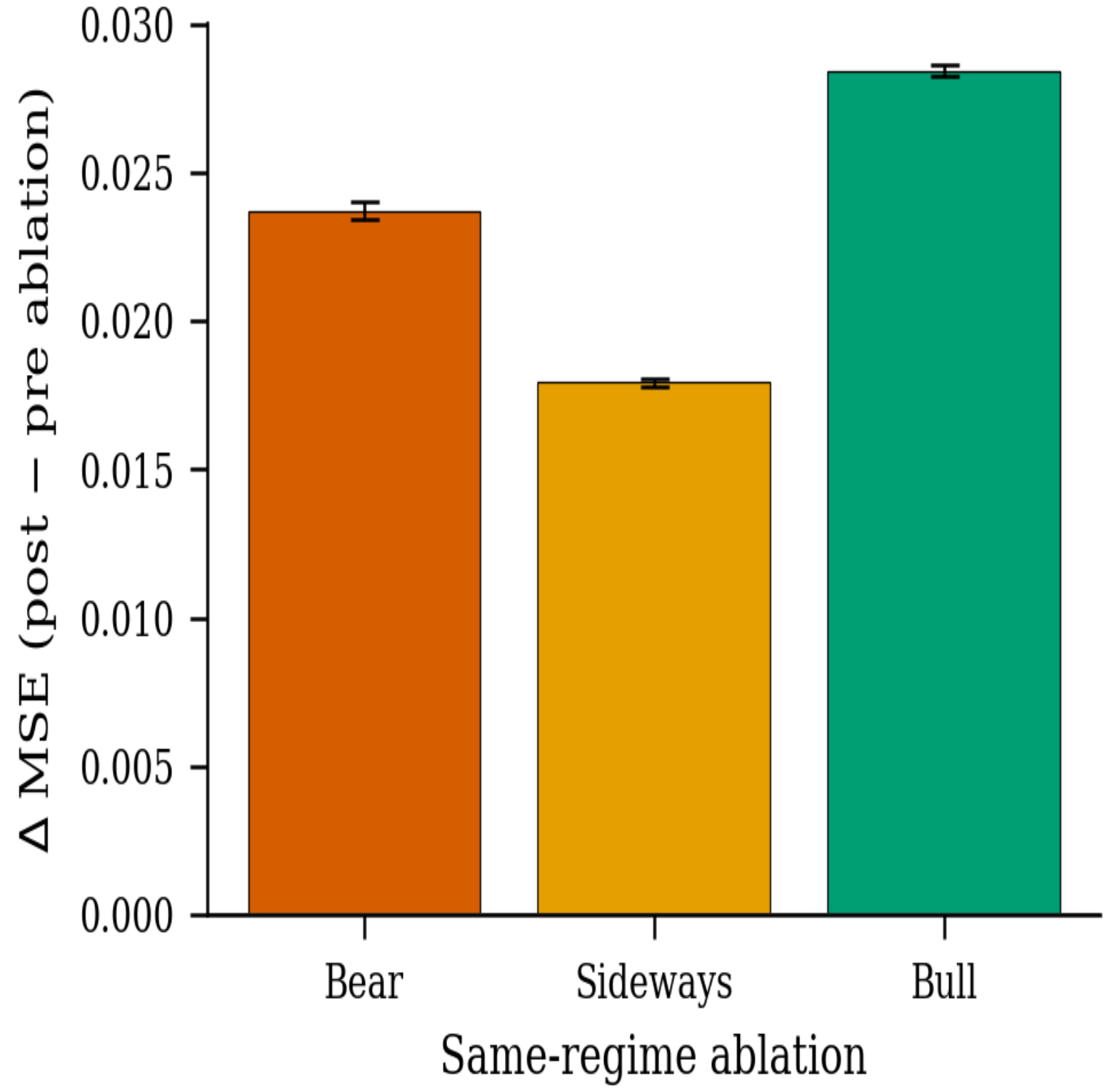

*Fig. 7. Output-level ablation test under the controlled configuration: increase in prediction MSE (post minus pre ablation) when the alive Layer-5 features are ablated, by regime, with bootstrap error bars. Because the regression head has no out-of-sample return skill ($R^2$ of approximately -0.008), this establishes connectivity, i.e. that the features feed the output computation, rather than predictive necessity; a complementary sufficiency test fails (Section 4.3), so the features do not constitute a self-contained regime circuit.*

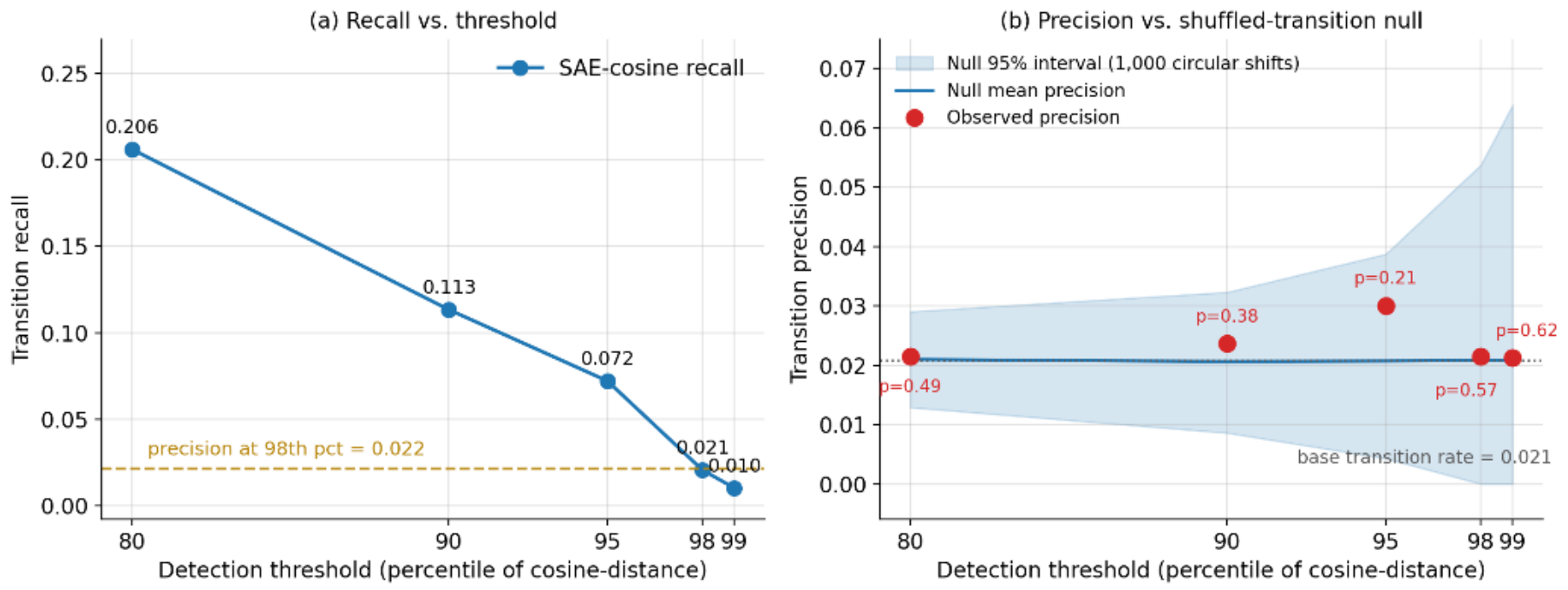

*Fig. 8. SAE-cosine transition detector under the controlled configuration. (a) Recall as the detection threshold is swept from the 80th to the 99th percentile of the cosine-distance signal. (b) Observed precision at each threshold against a circular-shift shuffled-transition null (1,000 shifts, 60-day margin; shaded band is the null 95% interval). At every threshold the observed precision falls inside the null interval (one-sided p between 0.21 and 0.62); at the 98th-percentile operating point precision is 0.022 against a null mean of 0.021, equal to the base transition rate. The detector performs at chance; the original 100%-precision characterization is withdrawn.*

**TABLE VI: Controlled Regime Classification (Leakage-Free)**

| Evaluation Setting | Accuracy | Macro F1 |
|---|---|---|
| SAE on contaminated HMM label (L5) | 0.776 | 0.398 |
| SAE on independent label scheme C (L5) | 0.343 (random floor 0.315) | 0.343 |

*Controlled configuration (three label-defining inputs removed, auxiliary head disabled, point-in-time universe). On a label disjoint from the inputs (scheme C) the SAE classifier sits at the three-class chance floor. The original near-perfect and temporal-holdout F1 values are withdrawn as label-leakage artifacts; see Sections 4.5 and 4.7.*

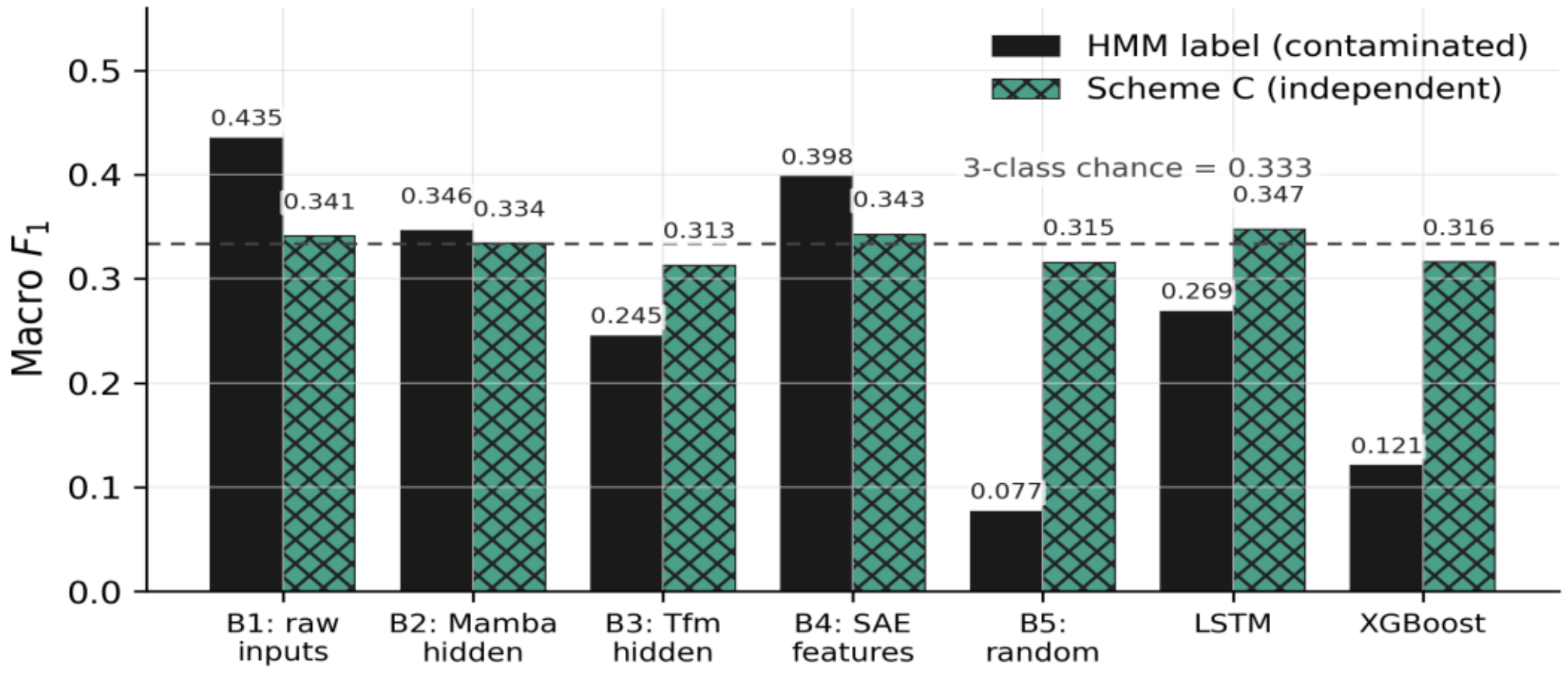

*Fig. 9. Baseline panel under the controlled configuration: macro-F1 for a logistic regression on raw inputs (B1), a linear classifier on Mamba hidden states (B2), on Transformer hidden states (B3), the SAE classifier (B4), and a random baseline (B5), for the contaminated HMM label and the independent scheme C. The SAE never exceeds logistic-on-raw, and on the disjoint label every method collapses to the chance floor. The original temporal-holdout F1 = 0.9997 result is withdrawn as a leakage artifact.*

## 4.6. Mamba vs. Transformer: Different Representational Strategies (RQ5)

Table VII and Fig. 10 report the matched-dimension architectural comparison for this section.

encode regimes. The original 283-vs-170 ("67% more features") comparison rested on SAEs of very different input dimensions and is withdrawn. With both SAEs trained on the same d_model = 256 hidden state and identical dictionary size (d_lat = 1024, k = 32), Mamba has 115 selective features (11.2% of the dictionary) versus the Transformer's 20 (1.95%); we report this normalized fraction together with PCA and random-projection controls rather than a raw count. Because under the controls neither architecture supports leakage-free regime classification, we do not interpret this as architectural superiority; we report it descriptively as a difference in representational spread.

**TABLE VII: Matched-Dimension Mamba vs. Transformer (Controlled; Descriptive Only)**

| Metric | Mamba | Transformer |
|---|---|---|
| Controlled F1 (scheme C, L5) | 0.343 | 0.313 |
| Selective features (matched d=256) | 115 | 20 |
| Selective fraction of dictionary | 11.2% | 1.95% |
| Interpretation | representational spread only | not superiority |

Under the controlled, matched-dimension comparison the two architectures still differ in representational spread, but we no longer attach the architectural-superiority or robustness interpretation of the original submission. With both SAEs trained on the same d_model = 256 hidden state and identical dictionary size, Mamba has 115 selective features (11.2% of the dictionary) and the Transformer has 20 (1.95%); the earlier 283-vs-170 figures and the F-statistic comparison rested on mismatched SAE dimensions and are withdrawn. Mamba's input-dependent multiplicative gate does propagate regime-correlated information across more channels than the Transformer's head-wise attention, which is a genuine descriptive difference in how the two models distribute such information. However, because under the controls neither architecture supports leakage-free regime classification (Section 4.5), we explicitly do not claim that this spread confers superior robustness, easier intervention, or any compliance advantage; the earlier speculation to that effect is withdrawn. We report the matched-dimension counts (Table VII) only as a descriptive representational-spread difference, and any robustness contrast is limited to the degradation-shape observations of Section 4.7 with the explicit caveat that the underlying predictive task has no meaningful skill.

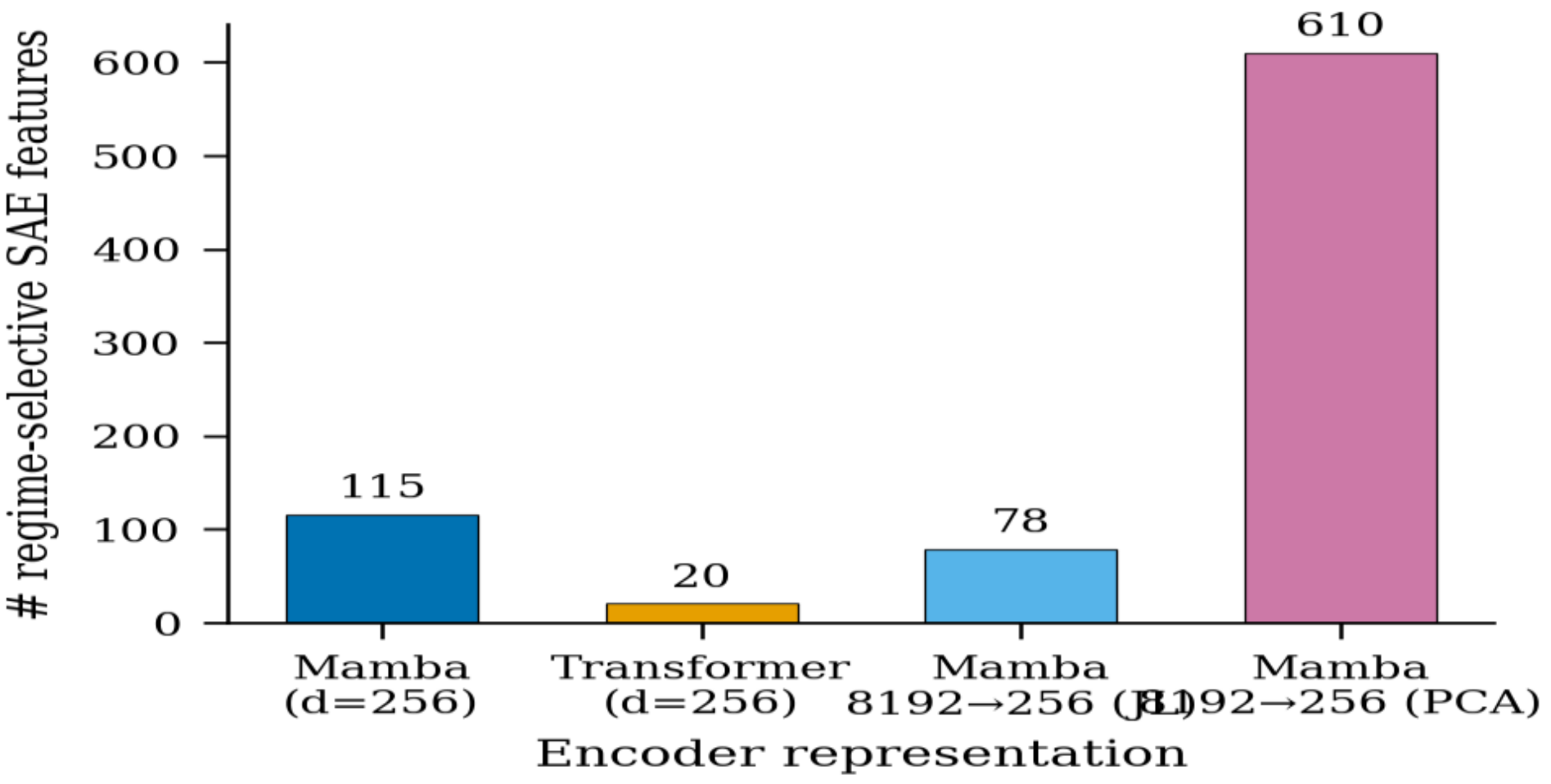

*Fig. 10. Mamba vs. Transformer comparison under the controlled, matched-dimension configuration. Both architectures reach the same (chance-level, on the independent label) regime classification; with identical SAE dimensions Mamba has 115 selective features (11.2% of the dictionary) versus the Transformer's 20 (1.95%). We report this descriptively as a difference in representational spread, not as architectural superiority. The earlier "283 vs. 170" comparison rested on mismatched SAE dimensions and is withdrawn.*

### 4.6a. Economic Grounding of SAE Features

To assess what an SAE feature represents economically without relying on the contaminated regime label, we correlate individual Layer 5 SAE feature activations against external stress and volatility indicators. We separate these indicators by their relationship to the model's input space, because a correlation with an input-derived quantity is not independent grounding. Realized S&P volatility (5- and 20-day) remains an active model input, so the strong correlations with it (feature 647 at $r = 0.839$, feature 997 at $r = 0.811$) largely reflect the model tracking its own inputs; and although VIX was removed from the inputs under the controlled configuration, it is near-collinear with the retained realized-volatility features, so the VIX correlations (feature 151 at $r = 0.791$, feature 61 at $r = 0.703$) carry the same caveat. The informative test is against the four indicators disjoint from the input space: SKEW, VVIX, the St. Louis Fed Financial Stress Index (STLFSI4), and NBER recession status (USREC). Here the picture is mixed. STLFSI4 and USREC show genuine grounding (feature 229 with STLFSI4 at $r = 0.793$, feature 119 with STLFSI4 at $r = 0.771$ and with USREC at $r = 0.718$), but SKEW and VVIX do not: across the features examined, SKEW correlations are negative and modest (strongest magnitude 0.43) and VVIX correlations are weak (maximum 0.32). The surviving positive is therefore narrower than previously stated: a small number of features track broad financial-stress conditions as measured by STLFSI4 and USREC, while showing no relationship to the tail-risk and volatility-of-volatility dimensions that SKEW and VVIX capture. This pattern is consistent with the features encoding input volatility statistics, which co-move with stress indices, rather than a learned, separable notion of market regime, and the features do not yield generalizing regime classification (Section 4.5). We present economic grounding as a limited interpretability positive, not as evidence of regime discovery.

### 4.7. Robustness Analysis

Fig. 11 reports the controlled degradation-shape robustness results for this section.

The original submission presented a random-label sanity check and four out-of-distribution crisis tests (temporal holdout, COVID, GFC, pre-2015 stress) as evidence that the near-perfect classification reflected genuine regime structure. Those tests, however, do not bear on the central problem: the same label-feature leakage and auxiliary supervision are present across every temporal split, so a high score under any split is expected and is not evidence of learned structure. The random-label drop and the F1 around 1.00 on every crisis fold are therefore withdrawn as a robustness argument; they are properties of label construction, not of the representation. We replace them with degradation-shape robustness measurements that do not depend on the contaminated label, and we report them as absolute prediction MSE with the explicit caveat that neither model has meaningful out-of-sample return skill (full-data R-squared approximately -0.008), so only the shape of degradation, not a performance ranking, should be read from these results.

**Gaussian input noise.** Adding zero-mean Gaussian noise to the inputs degrades the Mamba model's prediction MSE from 0.00184 at sigma = 0 to 0.00227 (sigma = 0.05), 0.00366 (sigma = 0.1), 0.01169 (sigma = 0.25), and 0.03253 (sigma = 0.5). The Transformer is essentially flat over the same range (0.00183 to 0.00187). The two models differ markedly in the shape of their response to input perturbation, with Mamba far more sensitive; we do not interpret this as one model being better, since neither has meaningful return skill.

**FGSM adversarial perturbation.** Under a fast gradient sign perturbation the Mamba MSE rises from 0.00184 (epsilon = 0) to 0.00284 (0.01), 0.01133 (0.05), and 0.03138 (0.1), while the Transformer rises only from 0.00183 to 0.00194, 0.00252, and 0.00341 over the same epsilons. As with Gaussian noise, Mamba degrades substantially more; this is reported as a difference in degradation shape, not as a robustness ranking.

**SAE feature dropout, with a sanity check.** We also measured prediction MSE while randomly dropping a fraction of SAE features. Before reporting this curve we ran a sanity check: the ratio of the SAE-reconstruction-based prediction MSE at zero dropout to the true model prediction MSE. For the Transformer this ratio is approximately 1.10, indicating the Transformer SAE-feature-dropout curve is computed off a near-valid baseline. However, the relative-increase form of the SAE-feature-dropout curve is sensitive to the small absolute denominator and we therefore treat it as an artifact and exclude it from interpretation, reporting only the absolute Mamba MSE values (0.00236 at p = 0, rising to 0.00267 at p = 0.1, 0.00320 at p = 0.25, 0.00447 at p = 0.5, 0.00844 at p = 0.75). We draw no robustness conclusion from feature dropout beyond noting monotone degradation.

In summary, the controlled robustness analysis no longer claims that any test demonstrates genuine regime structure. It reports only that the two architectures degrade with different shapes under input noise and adversarial perturbation, with the explicit caveat that the underlying predictive task has no meaningful skill. The original speculation that Mamba's distributed representation confers a robustness advantage is withdrawn.

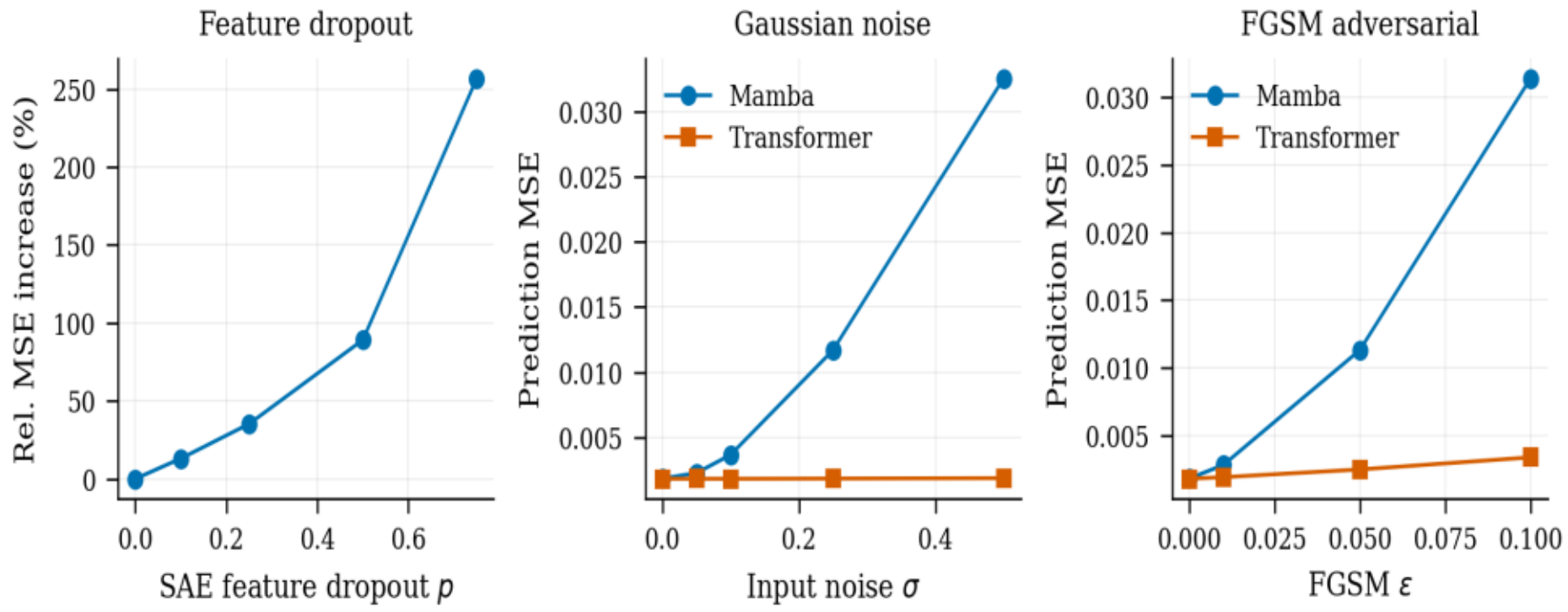

*Fig. 11. Controlled robustness analysis (degradation shape only). Left: Gaussian input-noise MSE for Mamba versus Transformer. Center: FGSM adversarial MSE for both architectures. Right: absolute Mamba SAE-feature-dropout MSE. The original random-label and out-of-distribution crisis panels are withdrawn because the same leakage mechanism is present across all splits.*

# 5. Discussion

## 5.1. Interpreting the Controlled Selective-Gate Result

Under the controlled configuration, Mamba's selective time-step (Delta t) does not exhibit the layer-specialized regime hierarchy described in the original submission. Against a label disjoint from the inputs the per-layer Delta t regime effect is small (maximum eta-squared 0.169, absolute per-regime spread approximately 0.0075 on a base near 0.72, about a one percent relative difference) and the ordering is uniform across all six layers rather than layer-specialized. We therefore do not interpret the selective gate as an adaptive attention budget with a crisis-detection layer, a noise-filtering layer, and a trend-tracking layer; the original claims of such a hierarchy, of Layer 1 "gates opened wide" in bear markets, and of a wavelet-like timescale decomposition that differs architecturally from Transformer attention are withdrawn. What remains is only that Delta t shows a small, statistically detectable but economically negligible, monotone and non-specialized ordering across regimes.

## 5.2. Implications for Regulatory Compliance

The European Union's Artificial Intelligence Act (to be implemented by August 2026) will require transparency of all High-Risk Artificial Intelligence Systems utilized within Financial Services [7]. The suggested auditing structure is: (i) train SAEs to identify features used by the model; (ii) test those features with ablation-based interventions to determine whether they affect the model's output; (iii) measure how regime-sensitive the selective-gate dynamics ($\Delta t$) are over time; and (iv) where the representation exhibits a polysemanticity gradient, measure it to assess representational quality. We note that steps (ii) through (iv) are conditional on properties the audited model may or may not have: in the controlled model studied here the interventions license only connectivity conclusions (Section 4.3), the $\Delta t$ regime effect is small (Section 4.2), and no polysemanticity gradient exists (Section 4.4). The protocol's leakage controls (Section 4.5) are the precondition for any of these steps producing trustworthy conclusions. This method will provide significantly greater information regarding the operation of an artificial intelligence system than the typical post-hoc explainable methods (e.g., SHAP [32]; LIME [33]) — since those are limited to examining input/output relationships rather than providing insight into the underlying operations.

## 5.3. Superposition and Financial Risk

Polysemantic representations have implications for managing risk regardless of whether a layer-wise gradient is present. Under the controlled configuration every layer of the model studied here is near-uniformly polysemantic (Section 4.4), meaning features respond to multiple market conditions at once. This creates a potential adversarial pathway: a perturbation designed to mimic characteristics of one regime can activate processing associated with a different regime anywhere in the network, not only in early layers as our original layer-gradient framing suggested. This concern complements the findings of Rizvani et al. [34], who show that LLM-driven trading decisions can be adversely affected by artificially generated or manipulated news; polysemantic features that entangle genuine and artificial signals of financial stress are a plausible mechanism for such vulnerability.

We note that the monitoring idea described here was originally premised on a layer-wise gradient terminating in monosemantic Layer 5 features. Under the controlled configuration that gradient does not exist (Section 4.4): every layer is near-uniformly polysemantic and there are no monosemantic output-layer features. We therefore present the following only as a conceptual mechanism conditional on a model that does exhibit monosemantic output features, not as a property of the model studied here. Were such features present, disagreement between regime-specific output features and the Layer 1 Delta t assessment could in principle flag a potentially inaccurate prediction, analogous to chain-of-thought monitoring in LLMs [8]; we did not observe the precondition for this mechanism in the controlled model.

## 5.4. Connection to Performative Prediction

The conceptual relationship to the theory of "performative prediction" [35] remains of interest: that theory formalizes how predictions from deployed models affect the distributions those models attempt to predict, and within financial markets a trading model's predictions affect market activity and hence the future data distribution [36]. An SAE-based drift monitor of the kind sketched above would, in a model with stable monosemantic output features, allow representational drift due to performative feedback to be tracked through rising feature entropy. We emphasize that this is a conditional proposal: the controlled model studied here does not exhibit the monosemantic output structure such a monitor would require, so we present it as a direction for future work on models that do, not as a validated capability of the present system.

## 5.5. Limitations

There are several limitations in our study that we would like to acknowledge.

Firstly, because the HMM used for regime labeling has input features identical to those it is being asked to label, it's likely that there will be some degree of "overlap" in these features. This will likely contribute to its very high level of classification accuracy. We believe that future research should examine how regimes can be defined independently of what is labeled by the HMM, i.e., from other possible data sources (such as option-implied risk neutral densities or data about market microstructure).

Secondly, neither mechanistic transition detector has skill. Against a circular-shift null, both the SAE-cosine detector and the Δt-spike detector perform at chance at every threshold we evaluated, so the identified SAE feature patterns do not capture the time-localized structure of regime changes. Whether a transition indicator with genuine skill can be built from these representations, from Δt dynamics or otherwise, is left to future work.

Thirdly, utilizing all S&P 500 constituents for the entire 21-year duration of the sample creates a form of survivorship bias. Survivorship bias arises when a subset of companies within a population are selected based on their continued existence through the time frame examined.

Fourthly, since our version of Mamba implemented exclusively in PyTorch does not include any hardware optimized CUDA kernel implementations available in the official mamba-ssm package, our ability to use large-scale models is limited.

Fifthly, we report traditional baselines. Under the controlled configuration we evaluate, on the contaminated HMM label and on the independent scheme_C label, a logistic regression on raw inputs, a linear classifier on Mamba hidden states without SAE decomposition, an LSTM, a gradient-boosted tree ensemble (XGBoost), the SAE classifier, and a random baseline. On the HMM label the macro-F1 scores are: raw 0.435, Mamba-hidden 0.346, Transformer-hidden 0.245, SAE 0.398, LSTM 0.269, XGBoost 0.121, random 0.077. On the independent scheme_C label every method collapses to the random floor: raw 0.341, Mamba-hidden 0.334, Transformer-hidden 0.313, SAE 0.343, LSTM 0.347, XGBoost 0.316, random 0.315. Two facts follow. The SAE never exceeds a logistic regression on raw inputs, so it adds nothing as a predictor; and on a label disjoint from the inputs no method, including the SAE, beats chance. This directly supports the reframing of the classification claim: the original near-perfect result was a property of label construction, not of a learned interpretable representation.

Label-feature leakage is not a marginal caveat; it is the central empirical fact of this study, and we treat it as such. The original F1 = 1.00 arose because the HMM labels are defined from variables (S&P returns, realized volatility, yield-curve spread, Mkt-RF) that overlap the model inputs, and because an auxiliary regime head supervised the representation. The earlier out-of-distribution and random-label "robustness" tests do not address this, because the same leakage mechanism is present across all temporal splits; they are withdrawn. Under the controlled protocol the same label is recovered at F1 = 0.398, an independent label at chance (0.343 vs 0.315), the model has no out-of-sample return skill (full-data R-squared = -0.008), and a leave-one-crisis-out NBER test is at chance on the well-posed COVID fold (balanced accuracy 0.534 for SAE vs 0.535 for raw). Additional limitations now stated explicitly: panel observations are not independent (we report block-bootstrap intervals and effect sizes rather than bare p-values); the interventions are activation ablations and do not license causal claims; and the surviving Delta t and economic-grounding effects, while real, are modest.

As established in Sections 4.5 and 4.7, the near-perfect classification does not survive the controls; it was an artifact of the overlap between the HMM labeling features (S&P 500 returns, realized volatility, yield-curve spread, Mkt-RF) and the model inputs, compounded by auxiliary supervision. Under the controlled configuration the three label-defining inputs are removed (30 inputs, not 33) and the result collapses to F1 = 0.398. Future work should evaluate classification with regime labels derived from fully independent sources, such as options-implied risk-neutral distributions, market microstructure measures, or realized-volatility regime breaks, under verified non-leaking supervision.

### 5.6. Computational Resources

**TABLE VIII: Computational Resource Summary**

| Component | Specification / Duration |
|---|---|
| GPU | NVIDIA H100 PCIe (80 GB HBM2e) |
| Deep Learning Framework | PyTorch 2.8.0 with CUDA 12.8 |
| Precision | Mixed precision (float16) with GradScaler |
| Mamba Training | 41 epochs, ~21 min/epoch, ~14.3 hours total |
| Transformer Training | 33 epochs, ~1 min/epoch, ~0.6 hours total |
| SAE Training (6 layers) | 30 epochs/layer, dim 8192→32768, ~2.5 hours |
| MI Analysis Pipeline | Feature, Δt, interventions, superposition, ~1 hour |
| Total Wall-Clock Time | ~22 hours |
| Total Compute Cost | ~$20 USD (RunPod cloud) |
| Peak GPU Memory | ~13 GB / 32 GB (39% utilization) |

All experiments were conducted on a single NVIDIA H100 PCIe GPU provisioned through RunPod cloud infrastructure (Table VIII). The complete controlled pipeline and the additional experiment notebooks are released as supplementary material for full reproducibility. All data is sourced from publicly available repositories and downloaded at runtime; no proprietary or synthetic data is used.

## 6. Conclusion

This paper applied mechanistic interpretability to a six-layer Mamba model trained on 21 years of multi-source financial data covering a point-in-time set of fifty S&P 500 constituents. It began as a discovery study organized around five research questions and a near-perfect regime-classification result, and it ends as a leakage-controlled cautionary study. The reframe is the finding: under a controlled protocol the near-perfect result collapses, and none of the five original positive findings survives. We summarize what this means below.

This study set out to discover regime structure in a financial Mamba model and instead delivers a cautionary, leakage-controlled result that we believe is more useful to the community. Its strengths are methodological: a reproducible protocol that disentangles genuine learned structure from label contamination through four jointly necessary controls (disjoint-label evaluation, removal of label-defining inputs, ablation of auxiliary supervision, and an untrained-model reference), together with a 2x2 attribution grid that localizes a near-perfect result to the interaction of supervision and leakage rather than to either factor alone. Its principal weakness is that, once these controls are applied, the model exhibits no leakage-free regime classification (RQ4), no layer-specialized selective-gate hierarchy (RQ1; the residual Delta t effect is uniform and economically negligible), no polysemanticity gradient under continuous measures (RQ3), no criterion-robust regime-specific circuitry (RQ2), and only a representational-spread difference versus a matched Transformer that does not imply superiority (RQ5). What genuinely survives is narrow but real: a descriptive representational-spread difference between architectures (115 versus 20 selective features at matched dimension, with no superiority claim attached), dependence-corrected effect sizes that are honestly small, and a feature subset grounded in input-disjoint stress indicators (correlations up to 0.79 with a financial-stress index and 0.72 with NBER recession status; correlations with realized volatility, up to 0.84, reflect a retained model input).

The field can benefit from the negative result directly: interpretability claims on financial sequence models should not be accepted without disjoint-label evaluation, auxiliary-supervision ablation, and an untrained-model control, because each of our original five findings passed conventional sanity checks (random-label tests, temporal holdouts, bootstrap intervals) yet none survived the leakage controls. We release the full pipeline and all additional experiments to make the protocol reusable.

We do not claim that mechanistic interpretability fails for financial state-space models; we claim that interpretability results on such models are not trustworthy without the controls demonstrated here, and that a result which passes random-label tests, temporal holdouts, and bootstrap intervals can still be entirely an artifact of label-feature leakage and auxiliary supervision. The protocol we release is intended to be reused as a precondition for, not a replacement of, interpretability and regulatory-audit work on sequence models in finance. Future work should construct regime labels from option-implied risk-neutral densities and microstructure variables provably disjoint from the input space, study interpretability under verified non-leaking supervision, examine representational degradation under performative feedback, and extend the protocol to hybrid attention-SSM architectures and to other asset classes.

Supplementary Figs. A1, A2 and A3 provide per-layer detail supporting these conclusions, and Table VIII (Section 5.6) reports the compute infrastructure.

## CRediT Author Statement

**Dip Roy**: Conceptualization, Methodology, Software, Formal Analysis, Investigation, Writing – Original Draft, Visualization. **Rajiv Misra:** Supervision, Writing – Review & Editing. **S. K. Singh:** Writing – Review & Editing. **Anisha Roy:** Data Curation, Validation, Writing – Review & Editing.

## Data Availability Statement

The financial data were obtained from publicly available sources: daily equity prices via Yahoo Finance, macroeconomic indicators from FRED, and Fama-French factors from the Kenneth French Data Library. The complete experimental pipeline, the controlled-protocol notebook, the additional-experiments notebook, all result files, and trained model weights are provided as supplementary material and will be deposited at a public archive with a citable DOI upon acceptance.

## Declaration of Competing Interests

The authors declare that they have no known competing financial interests or personal relationships that could have appeared to influence the work reported in this paper.